\PassOptionsToPackage{table,xcdraw}{xcolor}
\documentclass[11pt]{article}
\usepackage{acl}
\usepackage{times}
\usepackage{latexsym}
\usepackage[T1]{fontenc}
\usepackage[utf8]{inputenc}
\usepackage{microtype}
\usepackage{inconsolata}
\usepackage{graphicx}

\usepackage{tcolorbox}
\tcbuselibrary{skins}
\usepackage{fontawesome5}
\usepackage{xspace}  
\usepackage{multicol}
\usepackage{multirow}
\usepackage{booktabs}
\usepackage{pifont}
\definecolor{avggray}{gray}{0.93}
\usepackage{amssymb}
\usepackage{amsmath} 

\usepackage{tabularx}
\usepackage{array}

\definecolor{lightred}{RGB}{255, 230, 230}

\definecolor{mybg}{HTML}{FFFDE9}

\graphicspath{{img/}}   

\title{LatCom: Cross-Agent Latent Compression for Efficient Multi-Agent Collaboration}

\author{
\textbf{Shinan Zhang}$^{1}$\thanks{Equal contribution.},
\textbf{Tao Zhang}$^{1}$\footnotemark[1],
\textbf{Qihui Zhu}$^{1}$\footnotemark[1],
\textbf{Mengjie Zhang}$^{1}$,
\textbf{Dong Jin}$^{1}$,\\
\textbf{Yunpeng Hou}$^{2}$,
\textbf{Shuangwu Chen}$^{1}$\thanks{Corresponding author.},
\textbf{Xiaobin Tan}$^{1}$,
\textbf{Quan Zheng}$^{1}$,
\textbf{Jian Yang}$^{1}$\\
$^{1}$University of Science and Technology of China\\
$^{2}$Institute of Artificial Intelligence,
Hefei Comprehensive National Science Center\\
{\small
\texttt{
\{zsn884709682,zhangtaolqy,qh.zhu,zhangmengjie,kingdon,hyp314\}@mail.ustc.edu.cn
}}\\
{\small
\texttt{
\{chensw,xbtan,qzheng,jianyang\}@ustc.edu.cn
}}
}

\begin{document}
\maketitle
\begin{abstract}

LLM-based multi-agent systems (MAS) increasingly use latent collaboration to avoid the information loss and repeated encoding-decoding overhead of natural-language communication. However, directly forwarding all sender latents makes the receiver-side context scale with both the number of agents and the reasoning length, increasing computation, memory usage, and collaboration latency. A natural solution is latent compression. But we find that cross-agent redundancy remains unresolved in existing latent compression approaches, which typically compress each sender independently and then concatenate the results.
We propose \textbf{LatCom}, a cross-agent latent compression framework for efficient multi-agent latent collaboration. LatCom maps multiple sender latents into a fixed number of receiver-readable and task-relevant slots. Rather than reconstructing all sender hidden states, it optimizes the compressed latents for receiver-side task utility. LatCom trains the compressor in two stages: single-sender readability learning first establishes a latent interface interpretable by the frozen receiver, and multi-sender fusion learning then trains the compressor to fuse complementary evidence and remove redundancy across agents. Experiments on multiple benchmarks with Qwen3-4B show that LatCom achieves an average \(2.46\times\) inference speed-up over LatentMAS and reduces output token usage by \(70.3\%\), while maintaining comparable average accuracy.
\end{abstract}

\section{Introduction}
\label{sec:introduction} 
\vspace{-5pt}
LLM-based multi-agent systems (MAS) have emerged as a promising paradigm for complex reasoning, where specialized agents collaborate by exchanging intermediate thoughts and aggregating complementary evidence~\citep{
guo2024large,tran2025multi,zou2025latentmas}. 
\begin{figure}[t]
    \centering
    \includegraphics[width=1\linewidth]{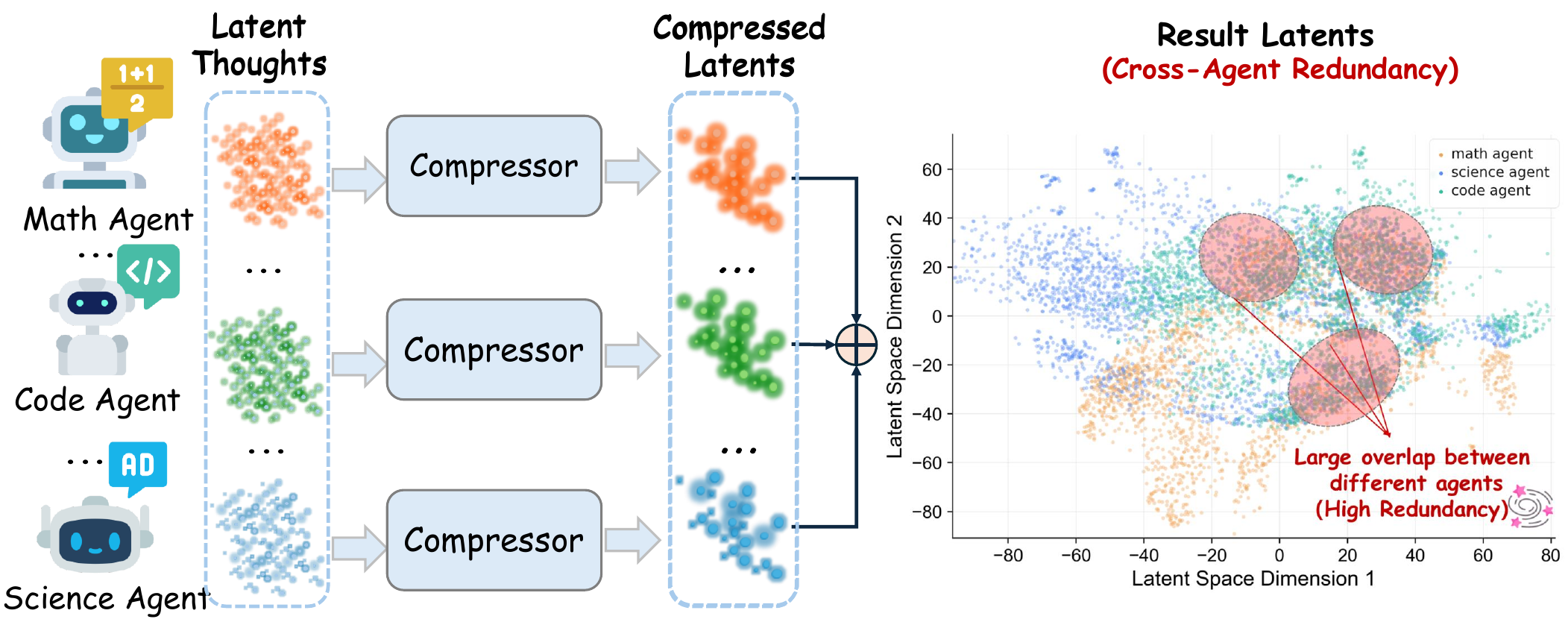}
    \caption{
    Independent per-agent compression preserves cross-agent redundancy, as t-SNE visualization shows overlap among concatenated compressed latents.
    }
    \label{fig:intro2}
    \vspace{-10pt}
\end{figure}
Most existing MAS communicate in natural language, requiring agents to project internal states into discrete token sequences~\citep{li2023camel,hong2024metagpt,wu2024autogen}. This discretization process may discard rich latent information and introduce additional overhead due to repeated encoding and decoding.~\citep{zhang2024cut,cemri2025multi,chen2025reasoning}. Latent collaboration addresses this limitation by allowing agents to exchange continuous internal representations, such as last-layer hidden states or KV caches, as intermediate latent thoughts or shared memory~\citep{hao2024training,zou2025latentmas,ramesh2025communicating,tang2025augmenting,zheng2025thought,du2025interlat,fu2025cache}.
However, latent collaboration introduces a new scaling bottleneck. As multiple agents generate latent thoughts over many reasoning steps, directly forwarding all latent thoughts makes the receiver-side latent context grow with both the number of agents and the reasoning length~\citep{zou2025latentmas,du2025interlat}. This increases receiver-side computation, expands memory usage, and slows the overall MAS collaboration process. 

A natural solution is to compress latent thoughts into compact and task-relevant representations~\citep{shen2025codi,cheng2024compressed}. Existing latent compression methods typically compress each sender's latent thoughts independently and then concatenate the compressed latents~\citep{du2025interlat} in the receiver side. 
However, as shown in Figure~\ref{fig:intro2}, by visualizing compressed latent thoughts of different agents via t-SNE~\citep{LatentMem}, we find substantial overlap even among highly compressed latent thoughts. 
Such redundancy arises from shared task context, overlapping evidence, and similar intermediate reasoning across agents. We further quantify this redundancy on GSM8K, ARC-Easy, and ARC-Challenge. Across the three datasets, pairwise raw cosine similarity ranges from \(0.553\) to \(0.771\), linear CKA ranges from \(0.408\) to \(0.600\), and the joint effective rank is approximately \(43\%\) lower than the sum of the individual ranks. Detailed results are reported in Appendix~\ref{app:cross_agent_redundancy}.This enables further latent compression to improve multi-agent collaboration efficiency.
%

Inspired by this finding, we propose LatCom, a cross-agent latent compression framework for efficient multi-agent latent collaboration. LatCom compresses multiple sender latents into a fixed number of receiver-readable and task-relevant latent slots. It trains the compressor in two stages: single-sender readability learning first establishes a latent interface interpretable by the frozen receiver, and multi-sender fusion learning then trains the compressor to fuse complementary evidence and remove redundancy across agents. Instead of reconstructing all sender hidden states, LatCom optimizes the compressed latents for receiver-side task utility. Experiments on multiple benchmarks with Qwen3-4B show that LatCom achieves an average \(2.46\times\) inference speed-up over LatentMAS and reduces output token usage by \(70.3\%\), while maintaining comparable average accuracy.

\textbf{Our Contributions}. 
\underline{(1) \textit{New Insight}.} We identify cross-agent latent redundancy as a key bottleneck in multi-agent latent collaboration, where independently compressed sender latents still contain duplication and increase receiver-side computation.
\underline{(2) \textit{New Framework}.} We propose LatCom, a cross-agent latent compression framework that maps multiple sender latents into a fixed number of  slots through two-stage compressor training.
\underline{(3) \textit{Comprehensive Evaluation}.} Extensive experiments on multiple LLMs and benchmarks  show that LatCom achieves strong task performance while substantially reducing inference latency and token usage.

\vspace{-4pt}

\section{Preliminary}
\subsection{Multi-Agent Latent Collaboration}
\label{sec:latent_collaboration}
We consider a multi-agent latent communication setting with sender agents \(\mathcal{A}_s=\{A_1,\ldots,A_N\}\) and a receiver \(A_R\)~\citep{zhao2026sirius,zhuge2024language}. Given task input \(q\), each sender \(A_i\) observes its context \(x_i\) and reasons directly in latent space. Instead of decoding intermediate thoughts into natural language, it emits aligned latent thought embeddings as continuous messages.
Let \(E_i=[e_{i,1},\ldots,e_{i,t}]\) denote the input embeddings of \(A_i\). At latent step \(\ell\), the sender maps its last-layer hidden state back to the input-embedding space:
\begin{equation}
z_{i,\ell}=h_{i,t+\ell-1}W_a^{(i)},
\end{equation}
where \(W_a^{(i)}\) is a sender-specific alignment operator~\citep{zou2025latentmas}. The aligned vector \(z_{i,\ell}\) is then reused as the next latent input, enabling iterative reasoning without intermediate text generation.
After \(m\) steps, sender \(A_i\) produces
\begin{equation}
    Z_i=\{z_{i,\ell}\}_{\ell=1}^{m}\in\mathbb{R}^{m\times d},
\end{equation}
where \(d\) is the hidden dimension. Latent communication avoids repeated decoding and re-encoding while preserving continuous reasoning signals. 
\subsection{Latent Compression}
\label{sec:latent_compression}
Forwarding all sender-side latent thoughts scales with both the number of senders \(N\) and latent steps \(m\). We therefore compress them into a fixed-size receiver-readable latent. Given sender trajectories \(\{Z_i\}_{i=1}^{N}\), we form a unified sequence
\begin{equation}
    U=[E_C(q),\rho_1,Z_1,\delta,\ldots,\rho_N,Z_N]\in\mathbb{R}^{L\times d},
\end{equation}
where \(E_C(q)\) embeds the task input, \(\rho_i\) marks sender identity, and \(\delta\) separates adjacent latents. This sequence preserves sender boundaries while exposing complementary evidence and redundancy to a latent compressor.
We instantiate \(C_\phi\) with \(K\) learnable slot tokens \(B=[b_1,\ldots,b_K]\in\mathbb{R}^{K\times d}\). These tokens collect information from the aggregated sender sequence. 
The compressor \(C_\phi\) maps \(U\) to a compact message 
\begin{equation}
    M=C_\phi(U,B)\in\mathbb{R}^{K\times d},
\end{equation}
where the \(K< L\).  
%
%
\begin{figure*}[t] 
  \centering
  \includegraphics[width=1\textwidth]{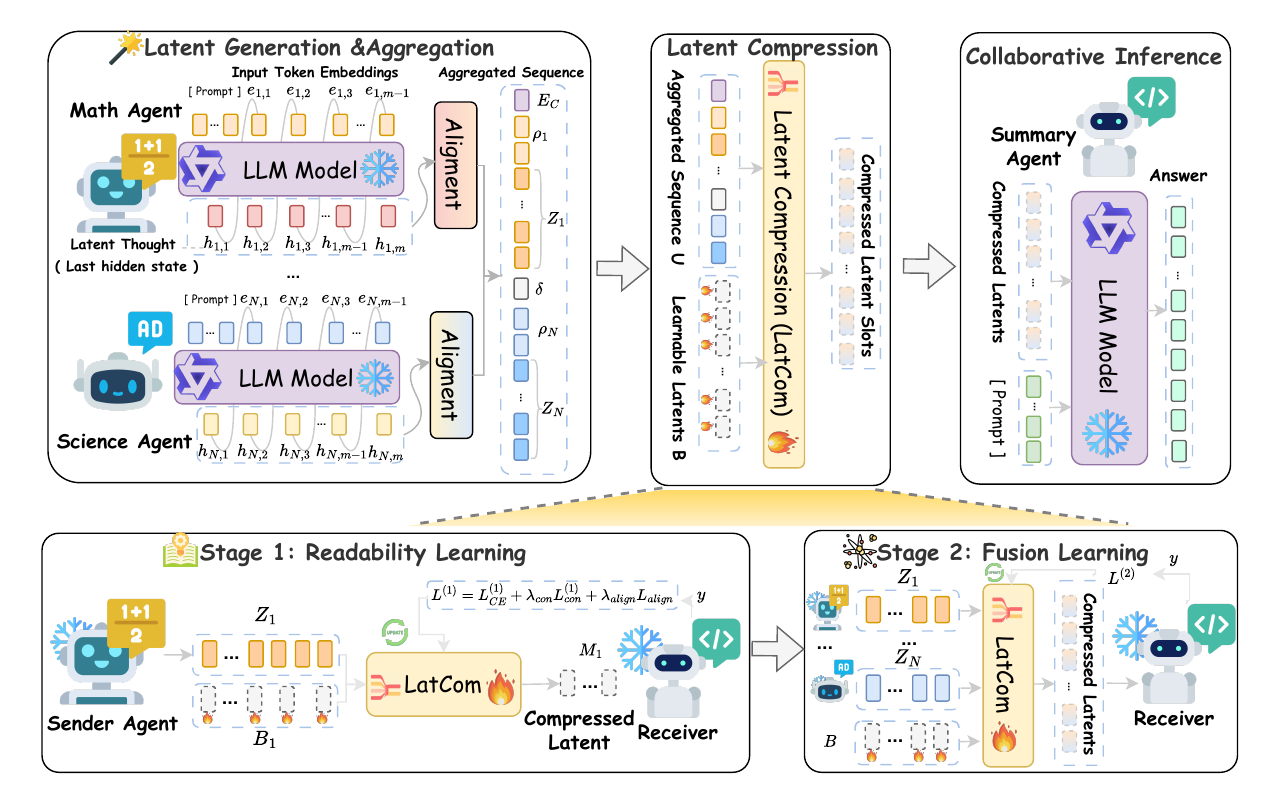}
    \caption{\textbf{Overview of LatCom.} Sender agents generate aligned latent thoughts, which LatCom aggregates and compresses into fixed-size receiver-readable slots. The frozen receiver uses these slots for final generation, while the compressor is trained through readability and fusion learning.}
  \label{fig:pipline}
\end{figure*}
Let \(E_R(q)\) denote the receiver-side prompt embeddings. The compressed slots are prepended to the receiver prompt as continuous context, and the receiver generates the output autoregressively:
\begin{equation}
p_{\theta_R}(y\mid q,M)
=
\prod_{t=1}^{|y|}
p_{\theta_R}
\bigl(
y_t
\mid
y_{<t},
[M,E_R(q)]
\bigr),
\label{eq:receiver_generation}
\end{equation}
where \(\theta_R\) are the receiver's parameters. Formally, latent compression is optimized by minimizing the frozen receiver's generation loss on the ground-truth answer \(y^\star\):
\begin{equation}
\phi^\star
=
\arg\min_{\phi}
\mathcal{L}
\bigl(
C_\phi(U), y^\star; \theta_R
\bigr),
\label{eq:compressor_objective}
\end{equation}
where \(\mathcal{L}(\cdot)\) evaluates how well the compressed latent message \(C_\phi(U)\) supports the receiver in generating \(y^\star\). The key challenge is to compress multi-sender latent thoughts without discarding task-critical information.

\section{LatCom}
\subsection{Overview of LatCom}
\label{sec:method}
Figure~\ref{fig:pipline} overviews LatCom, which consists of latent generation and aggregation, latent compression, and collaborative inference. Given a task input, sender agents produce aligned latent thoughts without decoding intermediate text. These thoughts are aggregated into a multi-sender sequence with task, sender-identity, and separator embeddings. LatCom then compresses the aggregated sequence into fixed-size receiver-readable latent slots. The compressor is trained in two stages: a single-sender readability stage that learns a receiver-compatible latent interface, followed by a multi-sender fusion stage that learns to compress complementary and redundant thoughts into shared slots. During inference, the compressed slots are prepended to the receiver prompt as continuous context, and the frozen receiver generates the final answer. 

\subsection{Two-stage Compressor Training}
\label{sec:two_stage_training}
We train LatCom in two stages. The compressor uses the same backbone family as the sender and receiver, with the language modeling head removed. During training, all sender and receiver parameters are frozen, and only the compressor is updated.
The two stages target different aspects of latent compression. Stage 1 uses a one-to-one setting, where a single sender's latent thought is compressed and consumed by the frozen receiver. This stage establishes a receiver-readable continuous interface. Stage 2 uses the final many-to-one setting, where multiple sender latent thoughts are jointly compressed into fixed-size slots. This stage trains the compressor to fuse task-relevant information while reducing the influence of irrelevant or redundant latent content.

\vspace{-3pt}
\paragraph{Evidence-structured training instances.}

LatCom is optimized for task-oriented compression rather than sender-state reconstruction.
To expose the compressor to varying source quality, we construct each training instance at the evidence level.
Evidence documents are grouped into gold-fact documents \(\mathcal{G}\), non-gold documents \(\mathcal{I}\), and mixed documents \(\mathcal{R}\) that combine partial gold and non-gold content.
These groups respectively produce task-supporting, irrelevant or distracting, and noisy but partially useful sender trajectories.

The two training stages use these evidence groups differently.
Stage 1 uses only gold-evidence inputs to learn a receiver-readable single-trajectory interface.
Stage 2 samples multiple sender inputs from all three groups, training the compressor to handle non-uniform source quality during multi-source compression.

\vspace{-3pt}
\paragraph{Receiver-side supervision.}
For any latent message \(M\), we define the receiver-side supervised decoding loss as
\begin{equation}
\small
\mathcal{J}(M)
=
-\frac{1}{|\mathcal{S}|}
\sum_{t\in\mathcal{S}}
\log p_{\theta_R}
\bigl(
y_t \mid y_{<t}, [M;E_R(q)]
\bigr),
\label{eq:receiver_loss}
\end{equation}
where \(\mathcal{S}\) indexes the supervised response tokens. Since the receiver is frozen, this loss trains the compressor through the receiver's generation behavior.

\vspace{-3pt}
\paragraph{Stage 1: Readability learning.}
Given a task input \(q\), target response \(y\), and one aligned sender trajectory \(Z_i\) from gold evidence, the compressor produces \(M_i=C_\phi(Z_i)\). The frozen receiver then predicts \(y\) conditioned on \([M_i;E_R(q)]\). Stage 1 first optimizes the receiver-side task loss:
\begin{equation}
\small
\mathcal{L}_{\mathrm{task}}^{(1)}
=
\mathcal{J}(M_i).
\label{eq:stage1_task_loss}
\end{equation}
To encourage the compressor to distinguish useful sender latents from unrelated ones, we introduce a receiver-grounded contrastive loss. 
For each matched thought \(Z_i\), we sample in-batch mismatched thoughts \(\mathcal{N}_i\). Let \(M^{-}=C_\phi(Z^{-})\) for \(Z^-\in\mathcal{N}_i\). We define
\begin{equation}
\small
\begin{aligned}
\mathcal{L}_{\mathrm{con}}^{(1)}
&=
\frac{1}{|\mathcal{N}_i|}
\sum_{Z^{-}\in\mathcal{N}_i}
\operatorname{softplus}(s_i^{-}),\\
s_i^{-}
&=
\frac{
\mathcal{J}(M_i)-\mathcal{J}(M^{-})+\Delta
}{\tau},
\end{aligned}
\label{eq:receiver_contrast}
\end{equation}
where \(\Delta\) is a margin and \(\tau\) is a temperature. This loss encourages the matched compressed latent to yield a lower receiver decoding loss than mismatched latents.
We further add a full-trajectory alignment loss to prevent the compressor from learning answer-triggering soft prompts that ignore sender-side information. 
Let \(P_t(X)\) denote the receiver predictive distribution over the token at position \(t\) conditioned on latent context \(X\):
\begin{equation}
    P_t(X)=p_{\theta_R}\bigl(\cdot\mid y_{<t},[X;E_R(q)]\bigr).
\end{equation}
Therefore, the alignment loss is defined as
\begin{equation}
\small
\mathcal{L}_{\mathrm{align}}^{(1)}
=
\frac{1}{|\mathcal{S}|}
\sum_{t\in\mathcal{S}}
D_{\mathrm{JS}}
\bigl(
P_t(Z_i), P_t(M_i)
\bigr)
+
\alpha
\left\|
\bar{M}_i-\bar{Z}_i
\right\|_2^2,
\label{eq:full_trajectory_alignment}
\end{equation}
where \(D_{\mathrm{JS}}\) is the Jensen-Shannon divergence, and \(\bar{M}_i\) and \(\bar{Z}_i\) are mean-pooled compressed and full sender latents.
The first term aligns receiver behavior under full and compressed latent contexts, while the second provides a coarse latent anchor.
The Stage 1 objective is therefore: 
\begin{equation}
\small
\mathcal{L}^{(1)}
=
\mathcal{L}_{\mathrm{task}}^{(1)}
+
\lambda_{\mathrm{con}}\mathcal{L}_{\mathrm{con}}^{(1)}
+
\lambda_{\mathrm{align}}\mathcal{L}_{\mathrm{align}}^{(1)},
\label{eq:stage1_objective}
\end{equation}
where \(\lambda_{\mathrm{con}}\) and \(\lambda_{\mathrm{align}}\) are weighting coefficients for the contrastive and alignment terms, respectively.

\vspace{-3pt}
\paragraph{Stage 2: Fusion learning.}
In stage 2, we train the LatCom in the final many-to-one setting. Each gold-evidence document \(\mathcal{G}\) is assigned to an individual sender, producing task-supporting latent thoughts from different sources. We further sample non-gold and mixed evidence from \(\mathcal{I}\) and \(\mathcal{R}\) with probabilities \(p_I\) and \(p_R\), respectively. The number of senders is sampled or clipped to the range of 2 to 6. The selected inputs are processed by frozen senders, aggregated into \(U\), and compressed into \(M=C_\phi(U)\).
The task loss is defined by the receiver-side decoding loss:
\begin{equation}
\small
\mathcal{L}_{\mathrm{task}}^{(2)}
=
\mathcal{J}(M).
\label{eq:stage2_task_loss}
\end{equation}
To train the compressor to identify task-supporting sources under non-uniform source quality, we extend the receiver-grounded contrast to multi-source inputs. The positive input is the matched aggregated sequence \(U\). Negatives \(\widetilde{U}\in\mathcal{N}_{\mathrm{ms}}(U)\) are constructed by replacing one or more gold-evidence sender inputs with same-type inputs from other examples. The corresponding messages \(M=C_\phi(U)\) and \(\widetilde{M}=C_\phi(\widetilde{U})\) are compared using the same receiver-loss margin contrast as in Eq.~\ref{eq:receiver_contrast}, yielding \(\mathcal{L}_{\mathrm{ms}}^{(2)}\).
Non-gold and mixed evidence sampled into the matched sequence \(U\) are not treated as negatives. They remain part of the input and expose the compressor to distracting, noisy, or redundant latent content. The contrastive term only penalizes cases where task-supporting sources are replaced by mismatched sources from other examples.
The Stage 2 objective combines the task and multi-source contrastive terms:
\begin{equation}
\small
\mathcal{L}^{(2)}
=
\mathcal{L}_{\mathrm{task}}^{(2)}
+
\lambda_{\mathrm{ms}}
\mathcal{L}_{\mathrm{ms}}^{(2)},
\label{eq:stage2_objective}
\end{equation}
where \(\lambda_{\mathrm{ms}}\) controls the strength of the multi-source contrastive loss.

\section{Experiments}
\label{sec:exp}
\subsection{Experimental Setup}
\label{sec:experimental_setup}

\begin{table*}[t]
\centering
\small
\setlength{\tabcolsep}{3.8pt}
\renewcommand{\arraystretch}{1.16}

\definecolor{LCUpText}{RGB}{36,128,62}
\definecolor{LCDownText}{RGB}{180,62,62}
\definecolor{LCSetting}{RGB}{238,238,238}
\definecolor{LCGSM}{RGB}{252,242,231}
\definecolor{LCARC}{RGB}{252,235,235}
\definecolor{LCMed}{RGB}{250,244,232}
\definecolor{LCCode}{RGB}{241,241,253}
\definecolor{LCGPQA}{RGB}{238,240,252}

\definecolor{LCAvg}{RGB}{232,241,250}
\definecolor{LCAvgText}{RGB}{45,68,95}

\newcommand{\LCbest}[1]{\textbf{#1}}
\newcommand{\LCup}[1]{\,{\scriptsize\textcolor{LCUpText}{($\uparrow$#1)}}}
\newcommand{\LCdown}[1]{\,{\scriptsize\textcolor{LCDownText}{($\downarrow$#1)}}}

\caption{\textbf{Main results of LatCom on 7 public benchmarks under the MAS setting.} Values in parentheses report the relative accuracy change of LatCom over each baseline. \textcolor{LCUpText}{$\uparrow$} and \textcolor{LCDownText}{$\downarrow$} indicate higher and lower accuracy, respectively.}
\label{tab:acc_results_vertical}

\resizebox{\textwidth}{!}{
\begin{tabular}{ll|cccccc}
\toprule
Tasks & Metrics
& TextMAS
& LatentMAS
& InterLat
& LatentMAS-H2O
& LatentMAS-hidden
& LatCom \\
\midrule

\multicolumn{8}{c}{\cellcolor{LCSetting}\textit{\textbf{Qwen3-4B}}} \\
\midrule

\cellcolor{LCGSM}GSM8K & Acc.
& 89.40\LCup{0.49\%}
& 88.10\LCup{1.98\%}
& 85.82\LCup{4.68\%}
& 83.55\LCup{7.53\%}
& 86.58\LCup{3.77\%}
& \LCbest{89.84} \\

\cellcolor{LCARC}ARC-E & Acc.
& 96.93\LCup{0.65\%}
& 95.45\LCup{2.21\%}
& 95.09\LCup{2.59\%}
& 94.74\LCup{2.98\%}
& 95.71\LCup{1.93\%}
& \LCbest{97.56} \\

\cellcolor{LCARC}ARC-C & Acc.
& 91.66\LCup{1.25\%}
& 91.72\LCup{1.19\%}
& 90.78\LCup{2.23\%}
& 89.85\LCup{3.29\%}
& 90.44\LCup{2.62\%}
& \LCbest{92.81} \\

\cellcolor{LCMed}MedQA & Acc.
& 65.49\LCdown{1.25\%}
& 66.33\LCdown{2.50\%}
& 64.16\LCup{0.79\%}
& 62.00\LCup{4.31\%}
& \LCbest{66.67}\LCdown{3.00\%}
& 64.67 \\

\cellcolor{LCCode}MBPP+ & Acc.
& \LCbest{69.12}\LCdown{2.40\%}
& 66.67\LCup{1.18\%}
& 65.74\LCup{2.62\%}
& 64.81\LCup{4.09\%}
& 64.81\LCup{4.09\%}
& 67.46 \\

\cellcolor{LCCode}HumanEval+ & Acc.
& 75.87\LCdown{0.34\%}
& \LCbest{76.22}\LCdown{0.80\%}
& 73.47\LCup{2.91\%}
& 70.73\LCup{6.90\%}
& 75.00\LCup{0.81\%}
& 75.61 \\

\cellcolor{LCGPQA}GPQA-Diamond & Acc.
& 41.00\LCup{12.02\%}
& 46.97\LCdown{2.21\%}
& 45.47\LCup{1.01\%}
& 43.97\LCup{4.46\%}
& \LCbest{47.25}\LCdown{2.79\%}
& 45.93 \\

\midrule
\rowcolor{LCAvg}
\textcolor{LCAvgText}{\textbf{Avg.}} & Acc.
& 75.64\LCup{0.83\%}
& 75.92\LCup{0.46\%}
& 74.37\LCup{2.56\%}
& 72.81\LCup{4.75\%}
& 75.21\LCup{1.41\%}
& \LCbest{76.27} \\

\midrule
\multicolumn{8}{c}{\cellcolor{LCSetting}\textit{\textbf{Qwen3-8B}}} \\
\midrule

\cellcolor{LCGSM}GSM8K & Acc.
& 91.45\LCup{1.64\%}
& 92.11\LCup{0.91\%}
& 91.81\LCup{1.24\%}
& 91.51\LCup{1.57\%}
& 92.19\LCup{0.82\%}
& \LCbest{92.95} \\

\cellcolor{LCARC}ARC-E & Acc.
& \LCbest{98.61}\LCdown{0.72\%}
& 96.72\LCup{1.22\%}
& 96.81\LCup{1.13\%}
& 96.89\LCup{1.04\%}
& 96.89\LCup{1.04\%}
& 97.90 \\

\cellcolor{LCARC}ARC-C & Acc.
& 93.65\LCup{1.19\%}
& 93.26\LCup{1.61\%}
& 92.15\LCup{2.83\%}
& 91.04\LCup{4.09\%}
& 93.86\LCup{0.96\%}
& \LCbest{94.76} \\

\cellcolor{LCMed}MedQA & Acc.
& 76.87\LCup{0.69\%}
& 75.45\LCup{2.58\%}
& 74.53\LCup{3.84\%}
& 73.62\LCup{5.13\%}
& 76.11\LCup{1.69\%}
& \LCbest{77.40} \\

\cellcolor{LCCode}MBPP+ & Acc.
& 72.19\LCup{0.78\%}
& \LCbest{73.51}\LCdown{1.03\%}
& 72.18\LCup{0.79\%}
& 70.85\LCup{2.68\%}
& 73.26\LCdown{0.70\%}
& 72.75 \\

\cellcolor{LCCode}HumanEval+ & Acc.
& 76.85\LCup{2.07\%}
& 78.06\LCup{0.49\%}
& 74.64\LCup{5.09\%}
& 71.22\LCup{10.14\%}
& 77.17\LCup{1.65\%}
& \LCbest{78.44} \\

\cellcolor{LCGPQA}GPQA-Diamond & Acc.
& 44.80\LCup{4.62\%}
& 47.93\LCdown{2.21\%}
& 46.53\LCup{0.73\%}
& 45.13\LCup{3.86\%}
& \LCbest{47.96}\LCdown{2.27\%}
& 46.87 \\

\midrule
\rowcolor{LCAvg}
\textcolor{LCAvgText}{\textbf{Avg.}} & Acc.
& 79.20\LCup{1.20\%}
& 79.58\LCup{0.72\%}
& 78.38\LCup{2.26\%}
& 77.18\LCup{3.85\%}
& 79.63\LCup{0.65\%}
& \LCbest{80.15} \\

\bottomrule
\end{tabular}
}
\end{table*}
\paragraph{Training data and LatCom training.}
LatCom is trained on multi-hop QA datasets, including HotpotQA~\citep{yang2018hotpotqa} and MuSiQue-Ans~\citep{trivedi2022musique}.
We filter out single-model-solvable, and pre-compression-unanswerable examples to focus training on multi-source evidence compression rather than base QA ability.
Training follows the two-stage procedure described in Section~\ref{sec:method}: Stage 1 learns receiver-readable slots, and Stage 2 trains multi-source compression with 2 to 6 senders.
Sender and receiver parameters are frozen throughout training, and only the compressor is updated.
Additional data construction details are provided in the appendix~\ref{app:training_data_construction}.

\paragraph{Datasets.}
We evaluate LatCom on seven public benchmarks covering mathematical reasoning, scientific and medical QA, commonsense reasoning, and code generation: GSM8K~\citep{cobbe2021training}, GPQA-Diamond~\citep{rein2023gpqa}, MedQA~\citep{yang2025llm}, ARC-Easy, ARC-Challenge~\citep{clark2018think}, MBPP-Plus, and HumanEval-Plus~\citep{liu2023your}.
This suite evaluates whether fixed-slot latent compression preserves task-relevant information across diverse reasoning domains and output formats.

\paragraph{Models and baselines.}
We use Qwen3-4B-Base and Qwen3-8B-Base~\citep{yang2025qwen3} as backbone LLMs.
All methods are evaluated in the same hierarchical three-sender setting~\citep{zhuge2024language}, where math, science, and code agents communicate with a receiver.
Baselines include TextMAS, LatentMAS~\citep{zou2025latentmas}, LatentMAS-Hidden, Interlat~\citep{du2025interlat}, and LatentMAS-H2O~\citep{zhang2023h2o}, covering text communication, KV-cache relay, aligned hidden-state transfer, sender-side latent compression followed by concatenation, and H2O-style KV-cache pruning.

\paragraph{Implementation details.}
Following LatentMAS~\citep{zou2025latentmas}, we compute the realignment matrix once per run and use 40 latent reasoning steps per sender. LatCom compresses the three aligned sender trajectories into 64 latent slots, which are prepended to the receiver prompt embeddings. For InterLat, we retrain the compressor using the same data sources, backbone, training budget, and evaluation setting as LatCom, and independently compress each sender into approximately 21 slots, resulting in approximately 63 slots in total, comparable to LatCom's 64-slot budget. All methods use identical sender and receiver models, decoding settings, and task-specific maximum output lengths. We report task performance and end-to-end latency on 8 NVIDIA A800-80G GPUs.

\subsection{Main Results}

Table~\ref{tab:acc_results_vertical} compares LatCom with text-based and latent-communication baselines under the MAS setting.
LatCom achieves the highest average accuracy with both Qwen3-4B and Qwen3-8B, reaching 76.27 and 80.15, respectively.
It obtains the best result in 7 out of 14 task-model settings and consistently outperforms LatentMAS-H2O and InterLat.
LatCom surpasses LatentMAS and LatentMAS-Hidden on average under both backbone sizes, showing that a fixed-slot latent interface can maintain strong accuracy without forwarding all sender-side latent states.
Compared with InterLat, which independently compresses each sender before concatenating the compressed latents, LatCom improves average accuracy by 1.9 points on Qwen3-4B and 1.77 points on Qwen3-8B.
This highlights the benefit of joint multi-source compression, where cross-sender redundancy and complementarity are modeled before receiver conditioning.
The gains are consistent across model scales, suggesting that the compression strategy is not tied to a specific backbone size.
Although the average gains over LatentMAS are modest and LatCom is not the best method on every task, it maintains comparable task performance with a bounded communication budget while substantially reducing inference cost.

\begin{figure}[t]
    \centering
    \includegraphics[width=1\linewidth]{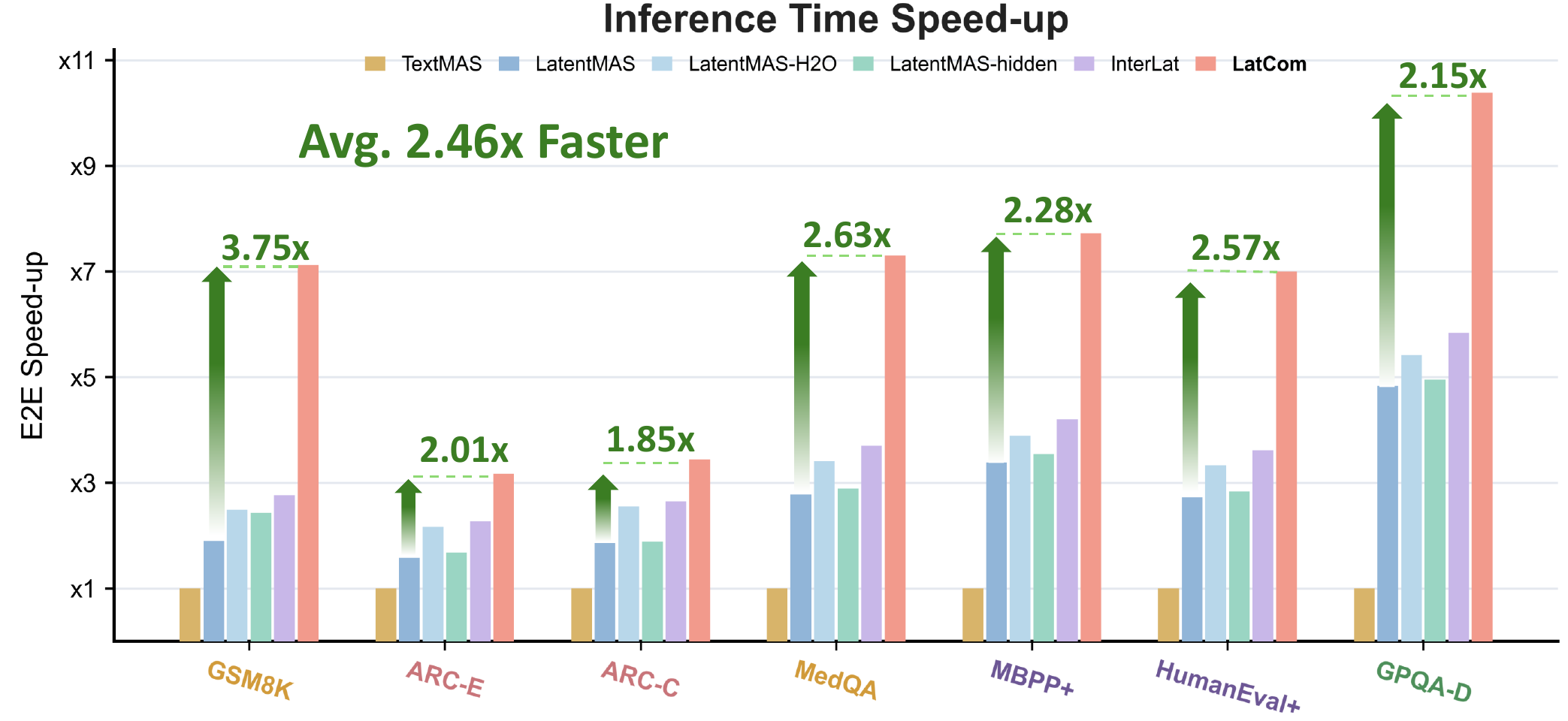}
    \caption{Inference speed-up across seven benchmarks.}
    \label{fig:e2e}
    \vspace{-10pt}
\end{figure}
\subsection{Efficiency Analysis}
We report end-to-end inference speed-up and average output token usage for each method on Qwen3-4B across multiple benchmarks.
As shown in Figure~\ref{fig:e2e}, LatCom achieves the fastest inference across all seven benchmarks, with an average speed-up of \(2.46\times\).
The gains are consistent across tasks, ranging from \(1.85\times\) on ARC-Challenge to \(3.75\times\) on GSM8K, indicating that the fixed-slot latent bottleneck reduces receiver-side processing cost.
We further profile the standalone compression step in Appendix~\ref{app:compressor_overhead}, showing that global compression takes only \(0.116\) seconds on average and therefore introduces limited runtime overhead.
Figure~\ref{fig:out-tokens} further shows that LatCom reduces output token usage by \(70.3\%\) on average, with task-level reductions ranging from \(63.6\%\) on ARC-Easy to \(81.3\%\) on GSM8K. This suggests that LatCom provides a compact conditioning signal that shortens receiver-side generation. Together with the accuracy results in Table~\ref{tab:acc_results_vertical}, these results show that task-oriented fixed-slot compression improves inference efficiency while preserving strong downstream performance.

\subsection{In-depth Analyses on LatCom}
\begin{figure}[t]
    \centering
    \includegraphics[width=1\linewidth]{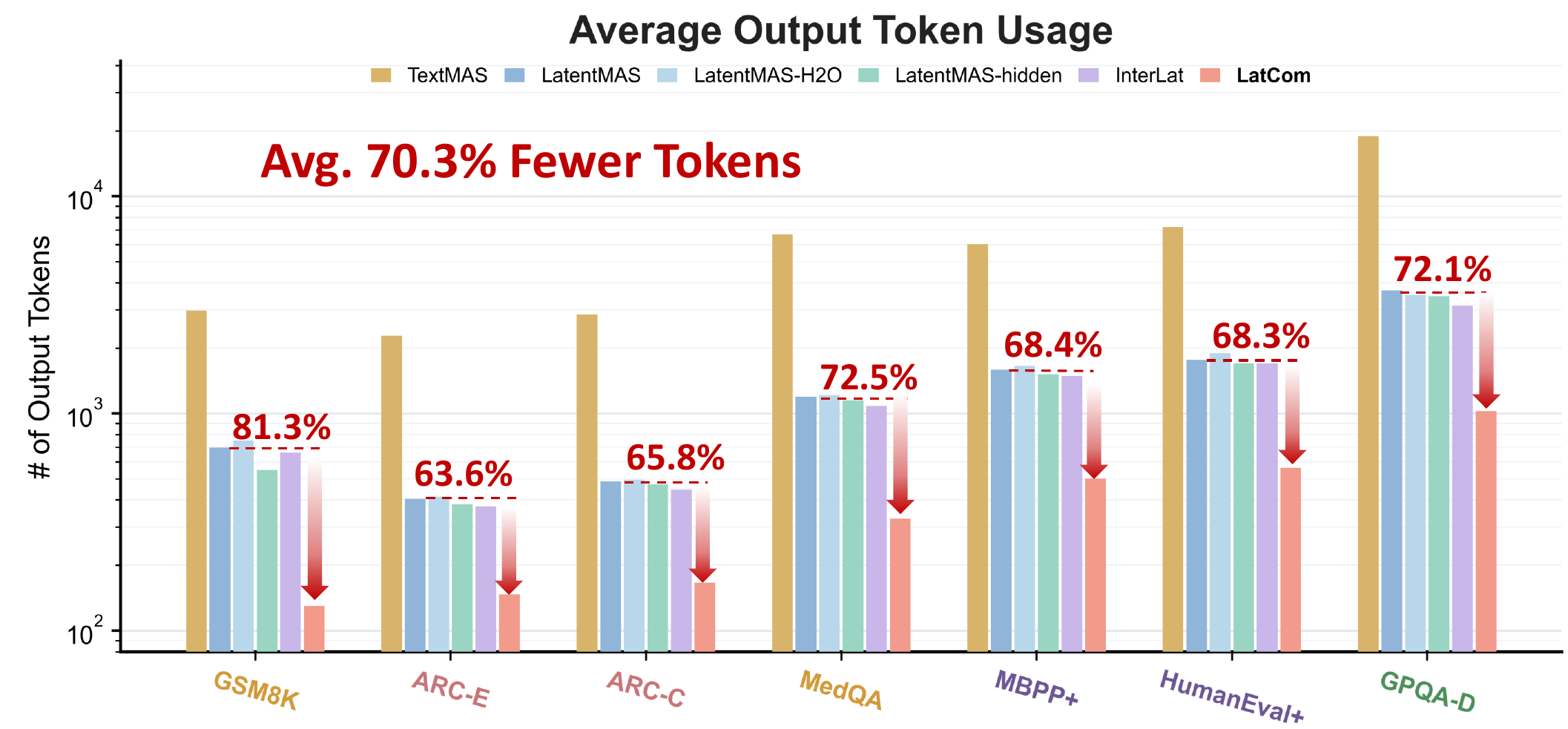}
    \caption{Output token usage across seven benchmarks. }
    \label{fig:out-tokens}
    \vspace{-10pt}
\end{figure}

\begin{figure}[t]
    \centering
    \includegraphics[width=1\linewidth]{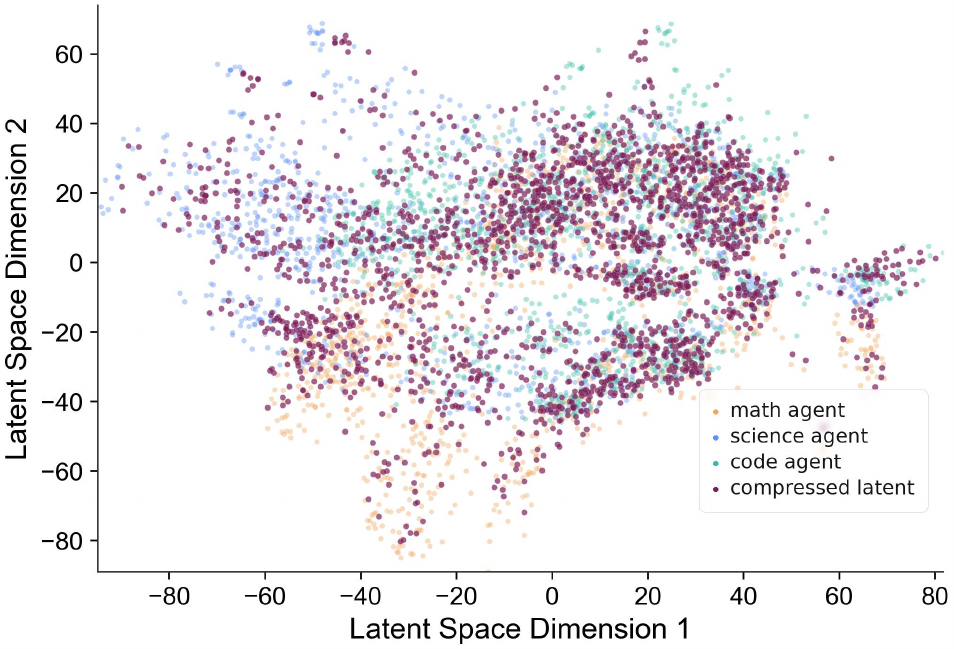}
    \caption{t-SNE visualization of latent communication}
    \label{fig:distribution}
    \vspace{-10pt}
\end{figure}

\paragraph{Does LatCom preserve useful latent structure after compression?}
We examine whether the compressed slots remain compatible with the sender-side latent space after fixed-slot compression.
Figure~\ref{fig:distribution} visualizes the aligned hidden states from three sender agents together with LatCom's compressed slots.
Sender hidden states occupy broad and highly overlapping regions, indicating semantically entangled and partially redundant latent trajectories.
By contrast, the compressed slots form a more concentrated distribution rather than reproducing the full sender-state space.
This suggests that LatCom learns a compact abstraction aligned with the sender space while filtering task-irrelevant or repeated latent variation.
The visualization provides qualitative evidence that fixed-slot compression preserves useful latent structure without directly forwarding all sender hidden states.

\begin{figure}[t]
    \centering
    \includegraphics[width=1\linewidth]{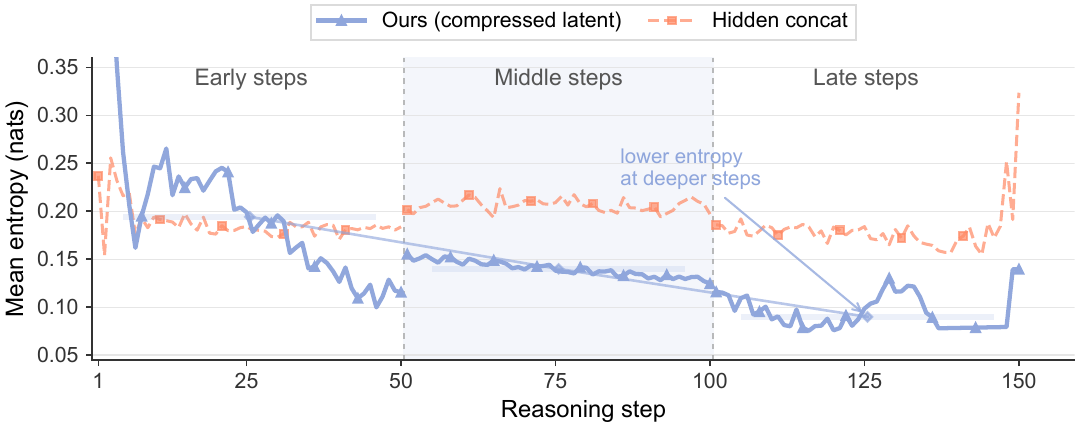}
    \caption{Receiver Entropy under Latent Conditioning}
    \label{fig:entropy}
    \vspace{-10pt}
\end{figure}

\paragraph{Does compression provide clearer task-oriented signals?}
We  further examine whether compressed slots provide a clearer conditioning signal for receiver-side generation. 
On GSM8K, we compare LatCom with LatentMAS-Hidden by measuring mean next-token entropy over generated reasoning traces, using three 50-token windows from the beginning, middle, and final parts of the reasoning body.
As shown in Figure~\ref{fig:entropy}, LatCom yields lower entropy across most reasoning positions, with a more stable gap in the middle and final windows.
This indicates that LatCom's compressed slots make the receiver's prediction distribution more concentrated, whereas direct conditioning on concatenated sender hidden trajectories may expose the receiver to redundant or weakly relevant latent signals that require additional decoding steps to filter and organize.
Together with the output-token reductions in Figure~\ref{fig:out-tokens} and the accuracy results in Table~\ref{tab:acc_results_vertical}, this suggests that LatCom provides cleaner receiver conditioning, enabling shorter generations while preserving competitive downstream performance.

\subsection{Ablation Study}
\begin{figure}[t]
    \centering
    \includegraphics[width=1\linewidth]{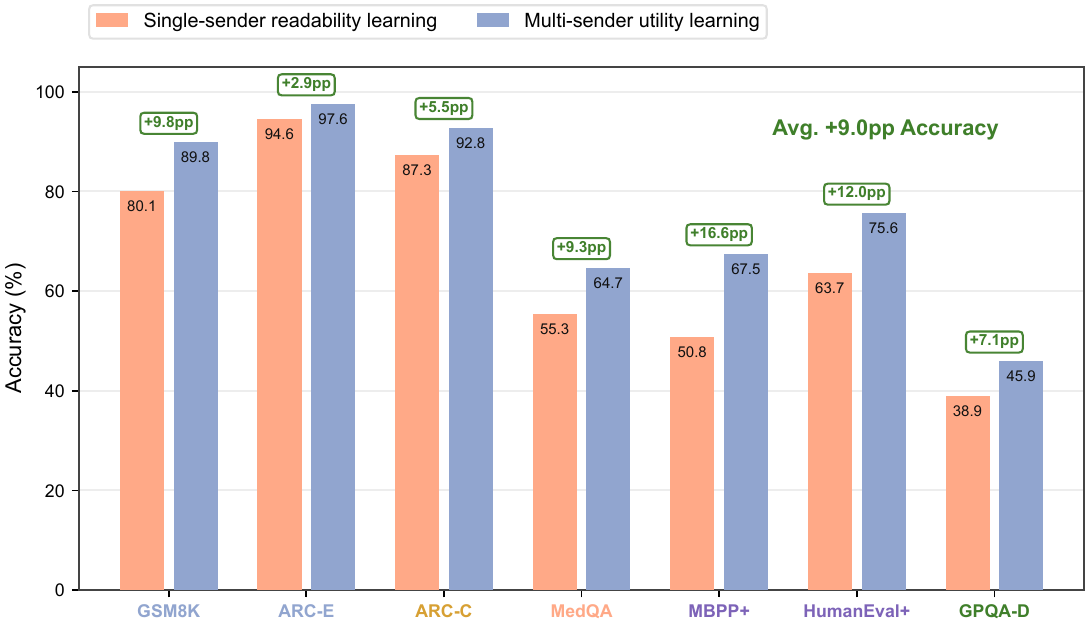}
    \caption{Ablation Study of Two-Stage Training}
    \label{fig:ablation_two_stage}
    \vspace{-10pt}
\end{figure}

\begin{table}[t]
\centering
\small
\setlength{\tabcolsep}{5.2pt}
\renewcommand{\arraystretch}{1.14}

\definecolor{LCStageFill}{RGB}{238,238,238}
\definecolor{LCFullFill}{RGB}{245,249,255}
\definecolor{LCDropText}{RGB}{180,62,62}
\definecolor{LCBestText}{RGB}{36,128,62}

\newcommand{\LCdrop}[1]{\textcolor{LCDropText}{\scriptsize\(\downarrow\)#1}}
\newcommand{\LCfull}[1]{\textbf{#1}}

\caption{\textbf{Ablation of training objectives on GSM8K.} ``Drop'' is
computed against the corresponding full setting within the same stage.}
\label{tab:ablation_components}

\begin{tabular}{llcc}
\toprule
Stage & Setting & Acc. & Drop \\
\midrule
\multicolumn{4}{c}{\cellcolor{LCStageFill}\textit{Single-sender readability learning}} \\
\midrule
\rowcolor{LCFullFill}
Stage 1
& Full Stage 1
& \LCfull{80.06}
& -- \\
Stage 1
& w/o \(\mathcal{L}_{\mathrm{con}}\)
& 74.53
& \LCdrop{5.53} \\
Stage 1
& w/o \(\mathcal{L}_{\mathrm{align}}\)
& 68.41
& \LCdrop{11.65} \\
\midrule
\multicolumn{4}{c}{\cellcolor{LCStageFill}\textit{Multi-source fusion learning}} \\
\midrule
\rowcolor{LCFullFill}
Stage 1 + Stage 2
& Full training
& \LCfull{89.84}
& -- \\
Stage 1 + Stage 2
& w/o \(\mathcal{L}_{\mathrm{ms}}\)
& 87.47
& \LCdrop{2.37} \\
\bottomrule
\end{tabular}
\end{table}

Figure~\ref{fig:ablation_two_stage} ablates the two-stage training procedure.
Single-sender readability learning produces receiver-readable slots, but remains limited to one-to-one communication.
Adding multi-sender utility learning improves accuracy on all benchmarks by an average of \(9.0\) percentage points, with larger gains on MBPP+ and HumanEval+ (\(16.6\) and \(12.0\) points).
This confirms that Stage 2 is important for adapting slots to useful multi-source compression.

Table~\ref{tab:ablation_components} further evaluates the training objectives on GSM8K.
Removing \(\mathcal{L}_{\mathrm{con}}\), \(\mathcal{L}_{\mathrm{align}}\), and \(\mathcal{L}_{\mathrm{ms}}\) reduces accuracy by \(5.53\), \(11.65\), and \(2.37\) points, respectively.
The largest drop from removing \(\mathcal{L}_{\mathrm{align}}\) shows that alignment to dense sender trajectories is critical for receiver-readable slots, while the other drops indicate that contrastive losses help bind slots to correct and task-supporting sources.

\subsection{Parameter Sensitivity Analysis}

\begin{figure}[t]
    \centering
    \includegraphics[width=1\linewidth]{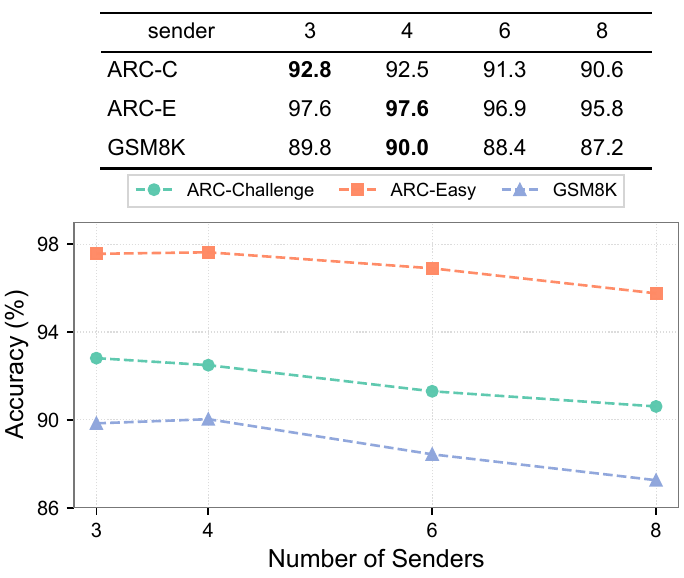}
    \caption{Sensitivity Analysis of Sender Number}
    \label{fig:sender_sensitivity}
    \vspace{-10pt}
\end{figure}

\paragraph{Effect of sender number.}
We examine LatCom under different numbers of senders with a fixed slot budget.
As shown in Figure~\ref{fig:sender_sensitivity}, performance is stable from 3 to 4 senders and declines moderately with 6 or 8 senders.
Even with 8 senders, the average drop on GSM8K, ARC-Easy, and ARC-Challenge is only about \(2.2\) percentage points, suggesting robustness to moderate sender scaling.

\begin{figure}[t]
    \centering
    \includegraphics[width=1\linewidth]{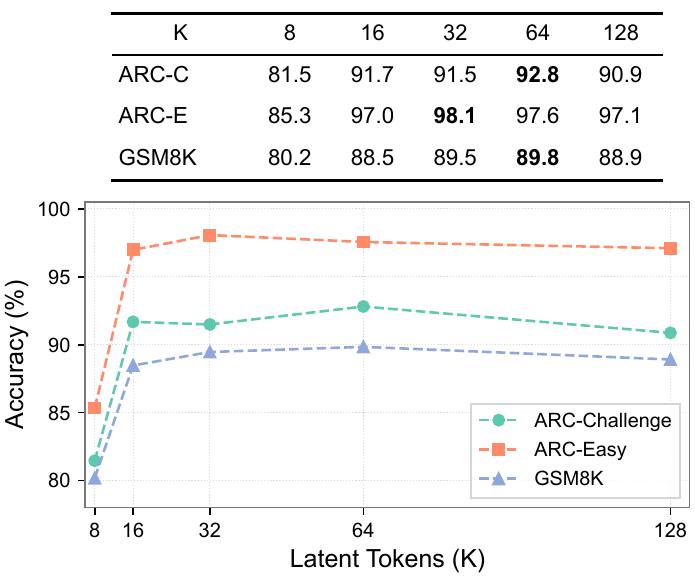}
    \caption{Sensitivity to Compressed Slot Number}
    \label{fig:compressed slot number sensitivity}
    \vspace{-10pt}
\end{figure}

\paragraph{Effect of compressed slot number.}
We further vary the compressed slot budget \(K\).
As shown in Figure~\ref{fig:compressed slot number sensitivity}, \(K=8\) causes clear accuracy drops, while performance improves and reaches the best or near-best accuracy around \(K=32\) to \(K=64\).
Since \(K=128\) brings no further gain and can slightly degrade performance, we use \(K=64\) in the main experiments as a stable accuracy-compactness trade-off.

\vspace{-4pt}
\section{Related Works}
\label{sec:related_work}
\vspace{-5pt}

\paragraph{LLM-based multi-agent systems.}
LLM-based multi-agent systems distribute problem solving across specialized agents, with representative frameworks including CAMEL~\citep{li2023camel}, MetaGPT~\citep{hong2024metagpt}, and AutoGen~\citep{wu2024autogen}.
Such systems can improve the coverage and robustness of complex reasoning~\citep{guo2024large,tran2025multi}, but most still rely on natural-language communication.
This requires agents to serialize internal states into tokens and downstream agents to re-encode them, which can discard fine-grained latent information and introduce additional inference overhead~\citep{zhang2024cut,cemri2025multi,chen2025reasoning}.
Our work addresses this communication bottleneck by studying compact latent representations for efficient inter-agent exchange.

\paragraph{Latent collaboration in multi-agent systems.}
Recent work explores continuous latent space as an alternative to text-mediated interaction.
Some studies perform latent reasoning within a single model, using hidden states instead of decoded chain-of-thought tokens~\citep{cheng2024compressed,chen2025reasoning}, while others extend latent interaction to model or agent collaboration through hidden-state exchange or KV-cache transfer~\citep{zheng2025thought,du2025interlat,fu2025cache,zou2025latentmas}.
However, existing latent-collaboration methods often relay full latent trajectories or directly combine states from multiple senders~\citep{zou2025latentmas,du2025interlat}, retaining redundant or low-utility states within trajectories and overlapping information across senders.
As the number of agents and reasoning steps increases, such unfiltered communication expands the receiver-side context and computational cost.
LatCom instead introduces a fixed-slot latent bottleneck for efficient multi-source exchange under a bounded communication budget.

\paragraph{Compression for multi-agent systems.}
Compression has been widely studied to improve LLM inference and communication efficiency.
Text-oriented methods, such as ICAE~\citep{ge2024incontext} and AutoCompressor~\citep{chevalier2023adapting}, compress natural-language context into soft memories but still rely on the text-to-latent conversion that latent collaboration seeks to avoid.
KV-cache reduction methods, including H2O~\citep{zhang2023h2o}, StreamingLLM~\citep{xiao2024efficient}, and SnapKV~\citep{li2024snapkv}, reduce long-context inference cost by retaining, evicting, or merging cached states, but are designed mainly for single-model cache management.
Closest to our work, hidden-state communication methods compress each sender's latent trajectory before receiver conditioning~\citep{du2025interlat}.
However, independent sender-side compression does not explicitly model cross-sender redundancy or complementarity before concatenation or aggregation.
LatCom instead performs receiver-aware multi-source compression, mapping multiple sender hidden trajectories into a fixed number of receiver-readable slots optimized for downstream task utility.

\section{Conclusion}
In this paper, we identified cross-agent latent redundancy as a key bottleneck in multi-agent latent collaboration and introduced LatCom, a cross-agent latent compression framework. LatCom compresses multiple sender latents into fixed-budget, receiver-readable, and task-relevant slots, mitigating receiver-side context growth. Its two-stage training establishes a latent interface for the frozen receiver and then learns to fuse complementary evidence while reducing cross-agent redundancy. Experiments across multiple benchmarks and models show that LatCom preserves strong task performance while substantially reducing inference latency and token usage, demonstrating the promise of latent compression for efficient multi-agent collaboration.

\section*{Limitations}
This work has several limitations.
First, LatCom relies on aligned hidden representations within the same model family.
As a result, its latent compression and receiver conditioning may be suboptimal in more general heterogeneous multi-agent systems, where agents differ in architecture, hidden dimensionality, tokenizers, or communication topologies.
Second, LatCom adopts a fixed-slot compression budget for multi-source latent trajectories.
Although this design improves efficiency, full latent relay remains stronger on some code-generation tasks, suggesting that fixed-slot compression may discard fine-grained information required for certain forms of reasoning.
Extending LatCom to heterogeneous model settings, adaptive communication budgets, and reliability-sensitive applications remains an important direction for future work.

\section*{Ethics Statement}
LatCom aims to improve the efficiency of multi-agent latent collaboration through task-oriented latent compression.
Because LatCom is built on existing LLMs, it may inherit their limitations, including biased predictions, hallucinations, and factually incorrect outputs.
Users should therefore verify model outputs when deploying such systems in real-world scenarios, especially in applications involving safety, privacy, or high-stakes decisions.
Our experiments are based on public benchmarks, Qwen3 models, PyTorch, and Hugging Face Transformers.
We follow their respective licenses and usage policies and gratefully acknowledge their contributions to the research community.

\section*{Acknowledgements}
This work was supported by the National Key R\&D Program of China,
No. 2024YDLN0004, and the Fundamental Research Funds for the Central
Universities under Grant No. WK2102026004.

\bibliography{custom}

\appendix
\section{Additional Analyses}

\subsection{Quantitative Analysis of Cross-Agent Redundancy}
\label{app:cross_agent_redundancy}

We quantify cross-agent redundancy on GSM8K, ARC-Easy, and ARC-Challenge using the latent trajectories produced by the math, science, and code agents. For each agent pair, we compute raw average cosine similarity, centered average cosine similarity, and linear centered kernel alignment (CKA). Raw average cosine measures the directional similarity between the original latent trajectories, while centered average cosine removes the mean direction before computing similarity.

Given two trajectory matrices \(X\) and \(Y\), linear CKA is computed as
\begin{equation}
\mathrm{CKA}(X,Y)
=
\frac{\lVert X^{\top}Y\rVert_F^2}
{\lVert X^{\top}X\rVert_F\,\lVert Y^{\top}Y\rVert_F},
\end{equation}
where \(\lVert\cdot\rVert_F\) denotes the Frobenius norm. Linear CKA measures the similarity between the overall representation structures of two agent trajectories.

\begin{table*}[t]
\centering
\small
\setlength{\tabcolsep}{7pt}
\renewcommand{\arraystretch}{1.1}
\begin{tabular}{llccc}
\toprule
Dataset & Agent pair & Raw AvgCos & Centered AvgCos & Linear CKA \\
\midrule
GSM8K & Math--Science & 0.605 & 0.061 & 0.600 \\
      & Math--Code & 0.693 & 0.061 & 0.559 \\
      & Science--Code & 0.771 & 0.069 & 0.508 \\
\midrule
ARC-Easy & Math--Science & 0.587 & 0.068 & 0.577 \\
         & Math--Code & 0.615 & 0.053 & 0.463 \\
         & Science--Code & 0.747 & 0.076 & 0.457 \\
\midrule
ARC-Challenge & Math--Science & 0.553 & 0.062 & 0.543 \\
              & Math--Code & 0.580 & 0.047 & 0.426 \\
              & Science--Code & 0.717 & 0.064 & 0.408 \\
\bottomrule
\end{tabular}
\caption{Pairwise similarity between latent trajectories from the math, science, and code agents.}
\label{tab:cross_agent_similarity}
\end{table*}

As shown in Table~\ref{tab:cross_agent_similarity}, raw average cosine similarity remains high across all three datasets. The Science--Code pair reaches 0.771, 0.747, and 0.717 on GSM8K, ARC-Easy, and ARC-Challenge, respectively. Linear CKA also remains consistently high; for example, the Math--Science pair obtains 0.600, 0.577, and 0.543. These results indicate that different agents share both common representational components and similar trajectory-level structures.

We further compute the effective rank of each agent trajectory and their joint trajectory. Given a hidden-state matrix \(H\) with singular values \(\{\sigma_i\}\), we define
\begin{equation}
\begin{aligned}
p_i
&=
\frac{\sigma_i^2}{\sum_j\sigma_j^2},\\
\mathrm{erank}(H)
&=
\exp\left(-\sum_i p_i\log p_i\right).
\end{aligned}
\end{equation}
For the three agents, \(\mathrm{SeparateSum}\) denotes the sum of their individual effective ranks. The joint trajectory is formed as
\begin{equation}
H_{\mathrm{Joint}}
=
[H_{\mathrm{Math}};H_{\mathrm{Science}};H_{\mathrm{Code}}].
\end{equation}
We measure the relative difference between the separate and joint effective ranks as
\begin{equation}
\mathrm{RankCompression}
=
1-
\frac{\mathrm{erank}(H_{\mathrm{Joint}})}
{\mathrm{SeparateSum}}.
\end{equation}

\begin{table*}[t]
\centering
\small
\setlength{\tabcolsep}{6pt}
\renewcommand{\arraystretch}{1.1}
\begin{tabular}{lrrrrrr}
\toprule
Dataset & Math & Science & Code & Separate Sum & Joint Rank & Rank Compression \\
\midrule
GSM8K & 7.6 & 10.3 & 9.6 & 27.5 & 15.6 & 43.3\% \\
ARC-Easy & 7.3 & 9.8 & 9.1 & 26.2 & 14.8 & 43.5\% \\
ARC-Challenge & 8.0 & 10.8 & 10.0 & 28.8 & 16.3 & 43.4\% \\
\bottomrule
\end{tabular}
\caption{Effective-rank analysis of individual and joint agent trajectories.}
\label{tab:cross_agent_effective_rank}
\end{table*}

Table~\ref{tab:cross_agent_effective_rank} shows that the joint effective rank is substantially smaller than the sum of the individual ranks. Rank compression remains close to 43\% on all three datasets, indicating that the effective dimensionality of the combined trajectories does not grow independently with the number of agents. This result is consistent with the similarity analysis and supports joint compression across sender trajectories.

\subsection{Controlled Analysis of Cross-Agent Fusion}
\label{app:fusion_controls}

We introduce two controls to separate the effect of joint multi-source fusion from simple pooling and additional single-source training. \textbf{Joint Mean Pool} takes the same aggregated multi-sender sequence as LatCom and applies adaptive mean pooling along the sequence dimension to produce 64 latent slots. It does not use a learned compressor. \textbf{Single-source Continuation} uses the same Transformer compressor, 64 learnable slots, training-data scale, and optimization budget as LatCom. It continues training on the examples allocated to Stage 2 while retaining the single-source construction, without forming joint multi-sender inputs.

\begin{table}[t]
\centering
\small
\setlength{\tabcolsep}{7pt}
\renewcommand{\arraystretch}{1.1}
\begin{tabular}{lcc}
\toprule
Method & Slots & GSM8K Acc. \\
\midrule
Joint Mean Pool & 64 & 78.92 \\
Single-source Continuation & 64 & 86.47 \\
LatCom & 64 & \textbf{89.84} \\
\bottomrule
\end{tabular}
\caption{Controlled comparison of cross-agent fusion under greedy decoding.}
\label{tab:fusion_controls}
\end{table}

As shown in Table~\ref{tab:fusion_controls}, Joint Mean Pool is 10.92 percentage points below LatCom, showing that simple pooling is insufficient for preserving task-relevant latent information. Single-source Continuation is 3.37 points below LatCom despite using the same compressor architecture, slot number, training-data scale, and optimization budget. The difference therefore comes from training on joint multi-sender inputs rather than from additional single-source training. This result is consistent with the 9.0-point average gain from Stage 2 and the 2.37-point drop after removing \(\mathcal{L}_{\mathrm{ms}}\).

\section{Training Details}
\label{app:training_details}

\subsection{Training Data Construction}
\label{app:training_data_construction}

We construct the main compressor training data from HotpotQA and MuSiQue-Ans.
HotpotQA is a Wikipedia-based multi-hop question answering dataset that provides questions, gold answers, passages, and sentence-level supporting-fact annotations.
Its questions often require bridge reasoning or comparison across multiple passages.
MuSiQue-Ans is a compositional multi-hop QA dataset in which questions require combining multiple single-hop reasoning steps, together with answer annotations and supporting evidence chains.

We use multi-hop QA as the primary training setting because it naturally requires information from multiple evidence sources.
Compared with single-hop factoid QA, this setting reduces the chance that the frozen receiver can solve the task from the question alone, making the training signal depend more directly on whether useful evidence is communicated through sender latents and preserved by the compressor.
To focus training on multi-source evidence compression rather than base QA ability, we filter out single-model-solvable examples and pre-compression-unanswerable examples.
The former can be answered by the backbone model without multi-source latent communication, whereas the latter cannot be answered even when the sender--receiver system receives the relevant sender information before compression.
The remaining examples therefore provide a cleaner signal for learning how to compress useful sender-side evidence into receiver-readable latent slots.

For each retained example, the task input is the question \(q\).
Because the original answers are often short spans or yes/no labels, answer-only targets provide limited receiver-side supervision.
We therefore construct a rationale-augmented target response from the question, supporting evidence, and gold answer.
Let \(\mathcal{E}^{+}\) denote the supporting evidence and \(a\) denote the gold answer.
A concise evidence-grounded rationale \(r\) is generated by an instruction-following model, and the target response is formatted as
\begin{equation}
\small
\begin{aligned}
r &= \mathrm{Rationale}(q,\mathcal{E}^{+},a),\\
y &= [\,\texttt{Reasoning: } r;\ \texttt{Answer: } a\,].
\end{aligned}
\label{eq:rationale_augmented_target}
\end{equation}
The rationale is constrained to remain consistent with the provided evidence, while the final answer follows the original dataset annotation.

\subsection{Evidence-structured Training Instances}
\label{app:evidence_structured_training}

We organize the associated passages according to their evidence annotations.
Documents containing supporting facts are treated as gold-evidence documents \(\mathcal{G}\), while the remaining documents are treated as non-gold documents \(\mathcal{I}\).
We additionally construct mixed evidence \(\mathcal{R}\) by combining partial gold-evidence content with partial non-gold content under the sender input-length budget:
\[
\begin{aligned}
\mathcal{G}&=\{g_j\}_{j=1}^{n_g},\quad
\mathcal{I}=\{i_j\}_{j=1}^{n_i},\\
\mathcal{R}&=\{r_j\}_{j=1}^{n_r}.
\end{aligned}
\]
These groups correspond to task-supporting, irrelevant or distracting, and noisy but partially useful sources.

Stage 1 uses only gold-evidence inputs.
Each instance contains the task input \(q\), the rationale-augmented target response \(y\), and a sender latent trajectory \(Z_i\) produced from gold evidence.
This stage focuses on learning a receiver-readable latent interface.

Stage 2 follows the final many-to-one communication setting.
Gold-evidence documents are assigned to different senders, and additional non-gold or mixed-evidence inputs are sampled to form heterogeneous multi-source inputs.
The number of senders is sampled or clipped to the range of 2 to 6.
To prevent the compressor from over-specializing to Wikipedia-style multi-hop QA, Stage 2 also includes a small number of auxiliary mathematical reasoning and code reasoning examples.
We convert these examples into the same multi-source format, with task-supporting, distracting, and redundant sources.
These auxiliary examples are used only to diversify the latent source structures seen during Stage 2 while preserving the same many-to-one compression objective.

The selected sender inputs are encoded by frozen senders, aggregated into \(U\), and compressed into fixed-size latent slots \(M=C_\phi(U)\).

\definecolor{LatComCaseTitle}{RGB}{190,112,35}
\definecolor{LatComCaseBack}{RGB}{255,250,243}
\definecolor{LatComCaseFrame}{RGB}{150,82,20}
\definecolor{LatComGold}{RGB}{241,248,239}
\definecolor{LatComNonGold}{RGB}{250,242,242}
\definecolor{LatComMixed}{RGB}{246,246,252}

\newtcolorbox{latcomcasebox}[1]{
  enhanced,
  colback=LatComCaseBack,
  colframe=LatComCaseFrame,
  coltitle=white,
  colbacktitle=LatComCaseTitle,
  title=\textbf{#1},
  fonttitle=\small\bfseries,
  fontupper=\footnotesize,
  boxrule=0.8pt,
  arc=2pt,
  left=5pt,
  right=5pt,
  top=4pt,
  bottom=4pt,
  width=\linewidth,
  before skip=4pt,
  after skip=4pt
}

\subsection{Example of Constructed Training Instance}
\label{app:constructed_training_example}

Figure~\ref{fig:constructed_training_example} illustrates how a raw multi-hop QA example is converted into evidence-structured sender inputs and a rationale-augmented target.

\begin{figure*}[t]
\centering
\begin{latcomcasebox}{Case Study: Evidence-structured Training Instance}

\textbf{Input Question.}
Which magazine was started first Arthur's Magazine or First for Women?

\vspace{2pt}
\textbf{Gold Answer.}
Arthur's Magazine

\tcblower

\textbf{Gold-evidence documents \(\mathcal{G}\).}
These passages contain the supporting facts needed to answer the question.

\vspace{2pt}
\begin{tcolorbox}[
  colback=LatComGold,
  colframe=LatComGold,
  boxrule=0pt,
  arc=1pt,
  left=4pt,
  right=4pt,
  top=3pt,
  bottom=3pt
]
\textbf{\(\mathcal{G}_1\): Arthur's Magazine.}
\textit{Arthur's Magazine (1844--1846) was an American literary periodical
published in Philadelphia in the 19th century.}
Edited by T.S. Arthur, it featured work by Edgar A. Poe, J.H. Ingraham,
Sarah Josepha Hale, Thomas G. Spear, and others. In May 1846 it was merged
into ``Godey's Lady's Book''.
\end{tcolorbox}

\vspace{2pt}
\begin{tcolorbox}[
  colback=LatComGold,
  colframe=LatComGold,
  boxrule=0pt,
  arc=1pt,
  left=4pt,
  right=4pt,
  top=3pt,
  bottom=3pt
]
\textbf{\(\mathcal{G}_2\): First for Women.}
\textit{First for Women is a woman's magazine published by Bauer Media Group
in the USA.}
\textit{The magazine was started in 1989.}
It is based in Englewood Cliffs, New Jersey. In 2011 the circulation of the
magazine was 1,310,696 copies.
\end{tcolorbox}

\vspace{3pt}
\textbf{Non-gold evidence documents \(\mathcal{I}\).}
These passages are fluent and partially related to temporal comparison, but
they do not provide the supporting facts required for the answer.

\vspace{2pt}
\begin{tcolorbox}[
  colback=LatComNonGold,
  colframe=LatComNonGold,
  boxrule=0pt,
  arc=1pt,
  left=4pt,
  right=4pt,
  top=3pt,
  bottom=3pt
]
\textbf{\(\mathcal{I}_1\): Echosmith.}
Echosmith is an American indie pop band formed in February 2009 in Chino,
California. Echosmith started first as ``Ready Set Go!'' until they signed to
Warner Bros. Records in May 2012.
\end{tcolorbox}

\vspace{2pt}
\begin{tcolorbox}[
  colback=LatComNonGold,
  colframe=LatComNonGold,
  boxrule=0pt,
  arc=1pt,
  left=4pt,
  right=4pt,
  top=3pt,
  bottom=3pt
]
\textbf{\(\mathcal{I}_2\): Freeway Complex Fire.}
The Freeway Complex Fire was a 2008 wildfire in the Santa Ana Canyon area of
Orange County, California. The ``Freeway Fire'' started first shortly after
9am with the ``Landfill Fire'' igniting approximately 2 hours later.
\end{tcolorbox}

\vspace{3pt}
\textbf{Mixed evidence \(\mathcal{R}\).}
Mixed evidence is constructed by combining partial gold evidence with partial
non-gold evidence under the sender input-length budget.

\vspace{2pt}
\begin{tcolorbox}[
  colback=LatComMixed,
  colframe=LatComMixed,
  boxrule=0pt,
  arc=1pt,
  left=4pt,
  right=4pt,
  top=3pt,
  bottom=3pt
]
\textbf{\(\mathcal{R}\): Example mixed source.}
Arthur's Magazine was published from 1844--1846. Echosmith is an American
indie pop band formed in February 2009. The Freeway Complex Fire was a 2008
wildfire in Orange County, California.
\end{tcolorbox}

\vspace{3pt}
\textbf{Rationale-augmented target.}
\begin{tcolorbox}[
  colback=white,
  colframe=black!18,
  boxrule=0.3pt,
  arc=1pt,
  left=4pt,
  right=4pt,
  top=3pt,
  bottom=3pt
]
\texttt{Reasoning:} Arthur's Magazine was published from 1844 to 1846.
First for Women was started in 1989. Since 1844 is earlier than 1989,
Arthur's Magazine was started first.

\vspace{2pt}
\texttt{Answer:} Arthur's Magazine
\end{tcolorbox}

\end{latcomcasebox}
\caption{Example of constructing evidence-structured sender inputs and a
rationale-augmented target from a HotpotQA instance.}
\label{fig:constructed_training_example}
\end{figure*}

\newcolumntype{Y}{>{\raggedright\arraybackslash}X}
\definecolor{LCGroupFill}{RGB}{238,238,238}

\begin{table*}[t]
\centering
\small
\setlength{\tabcolsep}{5.5pt}
\renewcommand{\arraystretch}{1.12}
\caption{Optimization and loss hyperparameters for LatCom training.}
\label{tab:latcom_optimization_hparams}
\begin{tabularx}{\textwidth}{p{0.20\textwidth} p{0.28\textwidth} Y}
\toprule
Category & Hyperparameter & Value \\
\midrule

\rowcolor{LCGroupFill}
\multicolumn{3}{l}{\textit{Optimization settings}} \\
Optimizer
& Optimizer type
& AdamW, \(\beta_1=0.9\), \(\beta_2=0.95\), \(\epsilon=10^{-8}\) \\

Optimizer
& Stage 1 learning rate
& \(2\times10^{-5}\) \\

Optimizer
& Stage 2 learning rate
& \(1\times10^{-5}\) \\

Optimizer
& LR scheduler
& Cosine decay \\

Optimizer
& Warmup steps / ratio
& \(5\%\) \\

Regularization
& Weight decay
& \(0.01\) \\

Batching
& Global batch size
& \(64\) \\

Batching
& Gradient accumulation steps
& \(4\) \\

Stability
& Max gradient norm
& \(1.0\) \\

\midrule
\rowcolor{LCGroupFill}
\multicolumn{3}{l}{\textit{Loss weights and contrastive settings}} \\
Task loss
& Stage 1 task loss weight
& \(1.0\) \\

Task loss
& Stage 2 task loss weight
& \(1.0\) \\

Contrastive loss
& \(\lambda_{\mathrm{con}}\)
& Dynamically adjusted in \([0.01, 0.5]\) \\

Alignment loss
& \(\lambda_{\mathrm{align}}\)
& Dynamically adjusted in \([0.01, 0.2]\) \\

Multi-source contrast
& \(\lambda_{\mathrm{ms}}\)
& Dynamically adjusted in \([0.01, 0.5]\) \\

Contrastive margin
& \(\Delta\)
& \(0.57\) \\

Contrastive temperature
& \(\tau\)
& \(1.0\) \\

Latent anchor
& Alignment anchor weight \(\alpha\)
& \(0.3\) \\

Supervision
& Supervised token set \(\mathcal{S}\)
& Rationale and answer tokens. \\

\bottomrule
\end{tabularx}
\end{table*}

\subsection{Training Hyperparameters}
\label{app:training_hyperparameters}

We report the optimization hyperparameters used for LatCom compressor training in Table~\ref{tab:latcom_optimization_hparams}.
Unless otherwise specified, sender and receiver parameters are frozen throughout training, and only the compressor parameters are updated.

\section{Evaluation Details}
\label{app:evaluation_details}

\subsection{Evaluation Benchmarks}
\label{app:evaluation_benchmarks}

We evaluate LatCom on the same seven-benchmark suite as LatentMAS, covering mathematical reasoning, scientific and medical question answering, commonsense reasoning, and code generation.
The suite includes GSM8K, ARC-Easy, ARC-Challenge, MedQA, MBPP-Plus, HumanEval-Plus, and GPQA-Diamond.

\paragraph{GSM8K.}
GSM8K is a grade-school mathematical reasoning benchmark consisting of natural-language word problems.
Each problem requires multi-step arithmetic reasoning and produces a final numerical answer.
We evaluate GSM8K by extracting and normalizing the final numeric prediction from the receiver output and comparing it with the gold answer.

\paragraph{ARC-Easy and ARC-Challenge.}
ARC-Easy and ARC-Challenge are multiple-choice science question answering benchmarks from the AI2 Reasoning Challenge.
ARC-Easy contains relatively straightforward elementary-level science questions, whereas ARC-Challenge contains more difficult questions that are less likely to be solved by shallow lexical cues.
We report multiple-choice accuracy after normalizing the predicted answer option.

\paragraph{MedQA.}
MedQA evaluates medical question answering in a multiple-choice format.
Compared with general commonsense QA, it requires domain-specific medical knowledge and careful interpretation of clinical or biomedical contexts.
We report accuracy by matching the normalized predicted option with the gold label.

\paragraph{GPQA-Diamond.}
GPQA-Diamond is a challenging graduate-level scientific question answering benchmark.
The Diamond subset contains expert-written questions designed to be difficult for non-experts and resistant to simple retrieval shortcuts.
We use it to evaluate whether latent compression preserves fine-grained scientific reasoning signals under a long-output setting, and report multiple-choice accuracy.

\paragraph{MBPP-Plus.}
MBPP-Plus is a code-generation benchmark extended from MBPP with stronger test cases.
Each example contains a natural-language programming problem, and the model must generate Python code that satisfies the benchmark tests.
We evaluate the generated program by executing it against the test cases and report pass rate.

\paragraph{HumanEval-Plus.}
HumanEval-Plus extends HumanEval with additional and more rigorous unit tests.
It evaluates functional code generation from problem descriptions and function signatures.
We extract the generated Python solution, execute it with the provided tests, and report pass rate.

\paragraph{Task-specific generation limits.}
Following the LatentMAS evaluation setting, we use task-specific maximum generation lengths.
GSM8K, ARC-Easy, and ARC-Challenge use a maximum output length of 2,048 tokens; MedQA, MBPP-Plus, and HumanEval-Plus use 4,096 tokens; and GPQA-Diamond uses 8,192 tokens.
These limits are shared across all compared methods.

\begin{table}[t]
\centering
\footnotesize
\setlength{\tabcolsep}{5.5pt}
\renewcommand{\arraystretch}{1.1}
\begin{tabular}{@{}llr@{}}
\toprule
Dataset & Category & Max. tokens \\
\midrule
GSM8K & Math & 2048 \\
ARC-Easy & Commonsense sci. & 2048 \\
ARC-Challenge & Commonsense sci. & 2048 \\
MedQA & Medical QA & 4096 \\
MBPP-Plus & Code & 4096 \\
HumanEval-Plus & Code & 4096 \\
GPQA-Diamond & Scientific QA & 8192 \\
\bottomrule
\end{tabular}
\caption{Evaluation benchmarks and maximum output lengths.}
\label{tab:evaluation_benchmarks}
\end{table}

\subsection{Compared Methods}
\label{app:compared_methods}

We compare LatCom with text-based and latent-communication baselines under the same hierarchical multi-agent setting.
Unless otherwise specified, all methods use the same sender roles, receiver role, prompts, decoding settings, and evaluation scripts.

\paragraph{TextMAS.}
TextMAS is the natural-language communication baseline.
Each sender agent generates a textual response according to its assigned role, and the receiver conditions on the concatenated sender outputs to produce the final answer.
This baseline represents the standard MAS communication interface, where intermediate information is serialized into discrete text before being used by the receiver.

\paragraph{LatentMAS.}
LatentMAS serves as the full latent-communication baseline.
It follows the hierarchical latent MAS setting, where sender agents perform latent rollout and transmit their latent computation states to the final receiver.
The receiver then generates the final answer conditioned on these latent states rather than relying only on natural-language messages.
In our experiments, LatentMAS uses the same hierarchical prompt structure, sender roles, latent reasoning steps, realignment setting, and decoding parameters as LatCom.

\paragraph{InterLat.}
InterLat is a learned latent-interface baseline for inter-agent communication.
Following its official setting, non-output agents exchange compressed continuous latent representations instead of natural-language messages, and the final output agent uses the received latent information to generate the answer.
We include InterLat as a representative latent-space communication method and evaluate it under the same benchmark suite and hierarchical comparison protocol.

\paragraph{LatentMAS-Hidden.}
LatentMAS-Hidden is a hidden-state relay variant of LatentMAS.
It follows the same hierarchical LatentMAS workflow, including sender and receiver roles, prompts, latent rollout steps, and realignment settings.
The key difference is that the sender-to-receiver relay object is changed from KV caches to hidden embeddings.
For each sender, we collect both the prompt/input hidden embeddings and the realigned latent hidden embeddings produced during latent rollout.
These embeddings are concatenated across senders and inserted into the final receiver's prompt embedding sequence, from which the receiver decodes the final answer.

\paragraph{LatentMAS-H2O.}
LatentMAS-H2O is a KV-compressed variant of LatentMAS.
It preserves the KV-cache relay mechanism of LatentMAS, but applies an H2O-style cache selection procedure before passing the sender cache to the final receiver.
During sender latent rollout, attention scores from latent tokens to previous positions are recorded and aggregated to estimate token importance.
The compressed cache retains important prompt tokens together with the preserved history and latent tail.
We use the headwise H2O variant as the default setting, with the same hierarchical prompt, latent rollout, realignment, and decoding setup as LatentMAS.

\begin{table*}[!t]
\centering
\small
\setlength{\tabcolsep}{5.2pt}
\renewcommand{\arraystretch}{1.12}
\begin{tabularx}{\textwidth}{@{}p{0.22\textwidth} p{0.28\textwidth} p{0.18\textwidth} Y@{}}
\toprule
Category & Metric & Value & Interpretation \\
\midrule

\rowcolor{LCGroupFill}
\multicolumn{4}{l}{\textit{Global compression time}} \\
Runtime & GSM8K & 0.095 s & Time for mapping \(U\) to \(M=C_\phi(U)\) \\
Runtime & ARC-Easy & 0.096 s & Time for mapping \(U\) to \(M=C_\phi(U)\) \\
Runtime & ARC-Challenge & 0.107 s & Time for mapping \(U\) to \(M=C_\phi(U)\) \\
Runtime & MedQA & 0.126 s & Time for mapping \(U\) to \(M=C_\phi(U)\) \\
Runtime & MBPP-Plus & 0.128 s & Time for mapping \(U\) to \(M=C_\phi(U)\) \\
Runtime & HumanEval-Plus & 0.132 s & Time for mapping \(U\) to \(M=C_\phi(U)\) \\
Runtime & GPQA-Diamond & 0.125 s & Time for mapping \(U\) to \(M=C_\phi(U)\) \\
\midrule
Runtime & Average & \textbf{0.116 s} & Average global compression time across tasks \\

\midrule
\rowcolor{LCGroupFill}
\multicolumn{4}{l}{\textit{Standalone compressor memory overhead}} \\
Memory & Compressor model footprint & 7.49 GiB & Persistent BF16 memory for loading one compressor \\
Memory & Peak CUDA allocation delta & 112.78 MiB & Additional peak GPU tensor memory during compression \\
Memory & Peak CPU RSS delta & 0.0086 MiB & Additional peak CPU memory during compression \\
\bottomrule
\end{tabularx}
\caption{Standalone overhead of the LatCom compressor. The model footprint is persistent, while CUDA and CPU deltas measure
the additional peak memory introduced by the compression call itself.}
\label{tab:compressor_overhead}
\end{table*}

\subsection{Implementation Details}
\label{app:implementation_details}

All methods are evaluated under a hierarchical multi-agent setting with three sender agents and one receiver agent.
Following the LatentMAS hierarchical setup, the senders are instantiated as math, science, and code agents, while the receiver serves as the final task summarizer.
The same hierarchical prompts are used for all compared methods.

For latent-communication methods, each sender performs \(40\) latent reasoning steps.
Latent-space realignment is enabled, and the realignment matrix is computed once per run.
Unless otherwise specified, all methods use greedy decoding with sampling disabled.
All experiments are implemented with HuggingFace Transformers and PyTorch, without vLLM.
Task-specific maximum output lengths follow Table~\ref{tab:evaluation_benchmarks}.

We use a unified answer extraction and evaluation protocol across methods.
For GSM8K, we extract the final numerical answer from the receiver output and compare it with the normalized gold answer.
For ARC-Easy, ARC-Challenge, MedQA, and GPQA-Diamond, we normalize the predicted option and compare it with the gold multiple-choice label.
For MBPP-Plus and HumanEval-Plus, we extract the generated Python code, combine it with the corresponding test cases, and execute the resulting program.
A prediction is counted as correct if it passes all benchmark tests within the execution timeout.

\subsection{Compressor Overhead}
\label{app:compressor_overhead}

We further measure the runtime overhead introduced by the LatCom compressor itself.
Specifically, we record the wall-clock time of the global compression step, namely the time required to map the aggregated multi-source latent sequence \(U\) into the fixed-slot message \(M=C_\phi(U)\).
This measurement excludes sender latent rollout, receiver prompt construction, and receiver decoding.

As shown in Table~\ref{tab:compressor_overhead}, global compression adds only a small computational overhead.
The average compression time is \(0.116\) seconds, indicating that the learned compressor introduces limited additional runtime relative to the overall multi-agent inference process.

\end{document}